\documentclass[letterpaper]{article} 
\usepackage[]{aaai2026}  
\usepackage{times}  
\usepackage{helvet}  
\usepackage{courier}  
\usepackage[hyphens]{url}  
\usepackage{graphicx} 
\usepackage{natbib}  
\usepackage{caption} 
\usepackage{algorithm}
\usepackage{algorithmic}

\usepackage{arydshln} 
\usepackage{booktabs}
\usepackage{graphicx}
\usepackage{amssymb,amsfonts,amsmath}
\usepackage{epsfig}
\usepackage{algorithm}
\usepackage{algorithmic}
\usepackage{booktabs}
\usepackage{mathrsfs}
\usepackage{color}
\usepackage{textcomp}
\usepackage{bbding}
\usepackage{pifont}
\usepackage{wasysym}
\usepackage{subcaption}
\usepackage{xcolor,colortbl}
\usepackage{overpic}
\usepackage{microtype}
\usepackage{multirow}
\usepackage{makecell}
\usepackage{hhline}
\usepackage[american]{babel}
\usepackage{cuted}
\usepackage{ragged2e}
\definecolor{lightgray}{gray}{.94}
\definecolor{tinygray}{gray}{.96}

\newcommand{\etal}{\textit{et al}.}

\newcommand{\eg}{\textit{e}.\textit{g}.}

\usepackage{multirow}
\usepackage{epsfig}
\usepackage{adjustbox}
\usepackage[pagebackref=true,
            breaklinks=true,colorlinks,
            citecolor=blue,linkcolor=red,
            bookmarks=false]{}

\usepackage{xcolor}
\newcounter{todos}
\AtEndDocument{\ifnum\value{todos}>0 \PackageWarning{TODOS}{There are \arabic{todos} todos left in this paper! Fix them before submitting the paper!} \fi}

\usepackage{newfloat}
\usepackage{listings}
\DeclareCaptionStyle{ruled}{labelfont=normalfont,labelsep=colon,strut=off} 
\floatstyle{ruled}
\newfloat{listing}{tb}{lst}{}
\floatname{listing}{Listing}
\title{Seeing Through the Rain: Resolving High-Frequency Conflicts in \\ Deraining and Super-Resolution via Diffusion Guidance}
\author{
    Wenjie Li\textsuperscript{\rm 1}
    Jinglei Shi\textsuperscript{\rm 2},
    Jin Han\textsuperscript{\rm 3},
    Heng Guo\textsuperscript{\rm 1}\thanks{Corresponding Author.},
    Zhanyu Ma\textsuperscript{\rm 1}
}
\affiliations{
    \textsuperscript{\rm 1}Beijing University of Posts and Telecommunications\\
    \textsuperscript{\rm 2}Nankai University
    \textsuperscript{\rm 3}The University of Tokyo\\

    \{cswjli, guoheng, mazhanyu\}@bupt.edu.cn
}

\usepackage{bibentry}

\begin{document}

\maketitle
\begin{abstract}
Clean images are crucial for visual tasks such as small object detection, especially at high resolutions. However, real-world images are often degraded by adverse weather, and weather restoration methods may sacrifice high-frequency details critical for analyzing small objects. A natural solution is to apply super-resolution (SR) after weather removal to recover both clarity and fine structures. However, simply cascading restoration and SR struggle to bridge their inherent conflict: removal aims to remove high-frequency weather-induced noise, while SR aims to hallucinate high-frequency textures from existing details, leading to inconsistent restoration contents. In this paper, we take deraining as a case study and propose DHGM, a Diffusion-based High-frequency Guided Model for generating clean and high-resolution images. DHGM integrates pre-trained diffusion priors with high-pass filters to simultaneously remove rain artifacts and enhance structural details. Extensive experiments demonstrate that DHGM achieves superior performance over existing methods, with lower costs. Code link: \url{https://github.com/PRIS-CV/DHGM}.
\end{abstract}

\section{Introduction}
\begin{figure}[t]
	\begin{overpic}[width=0.98\linewidth]{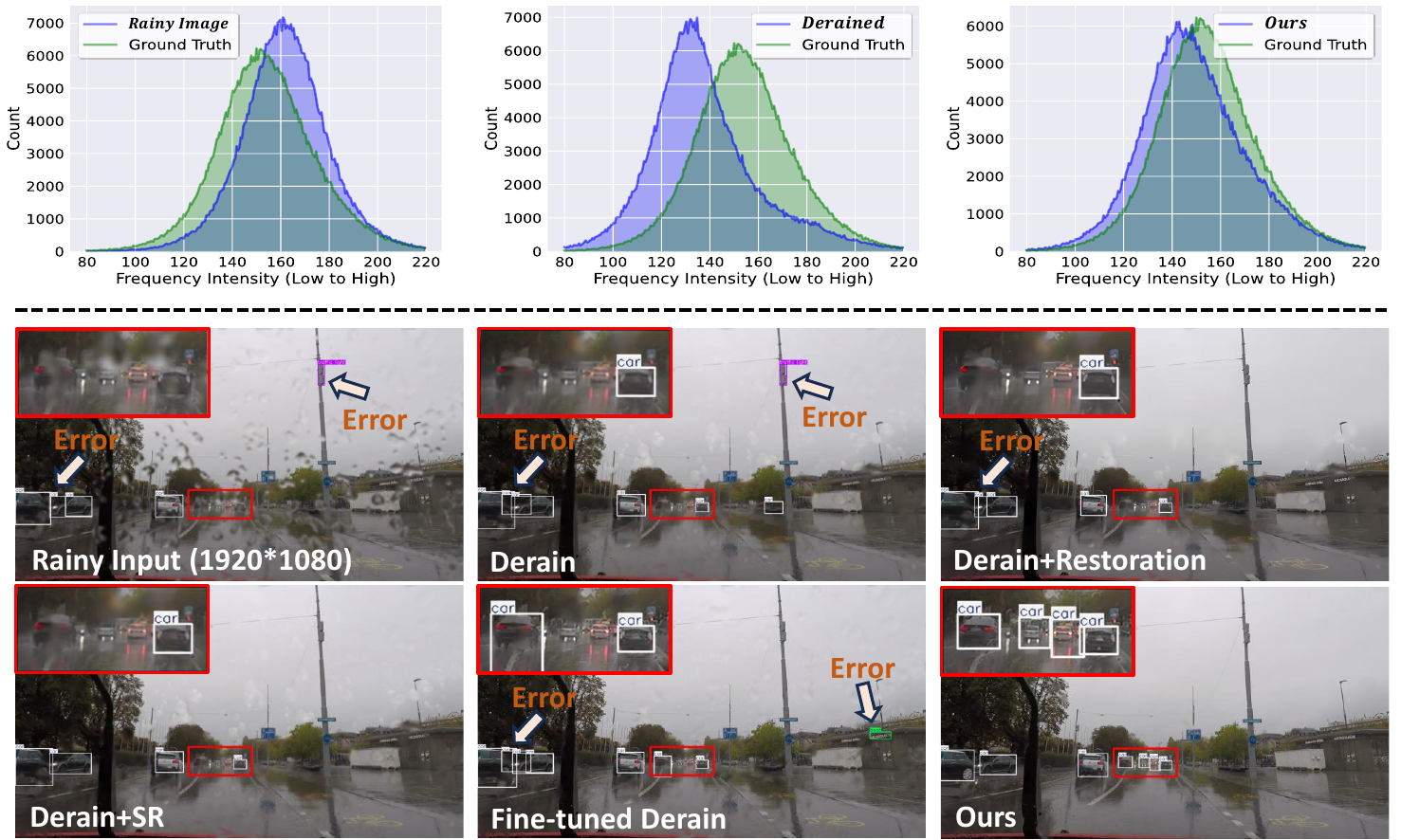}
	\end{overpic}
	\caption{(\textbf{Top}) Frequency-domain analysis shows that weather removal methods eliminate not only rain streaks but also valuable high-frequency textures. (\textbf{Bottom}) Compared with existing methods, our method better preserves details and improves small-object detection under rainy conditions.}
	\label{fig:teaser_small}
\end{figure}

    Adverse weather commonly degrades visual quality, significantly impacting downstream tasks such as object detection~\cite{varghese2024yolov8}. To mitigate these degradations, weather restoration is typically employed to remove weather-induced noise and recover clean images. However, as pointed out in recent studies~\cite{yang2017deep,jin2020ai}, existing methods inevitably sacrifice high-frequency details, resulting in excessive smoothing. This observation is also confirmed by our spectral visualization at the top of Fig.~\ref{fig:teaser_small}, showing that high-frequency parts in weather removal images are attenuated compared to ground truth (GT). This indicates that existing methods remove not only weather noise but also high-frequency textures. Such an issue severely impairs detection of small targets, such as distant vehicles, since even at 2K resolutions, these objects are low-resolution, typically occupy only a few dozen pixels, making accurate detection particularly sensitive to texture or edge loss introduced by weather removal.

To preserve and reconstruct essential high-frequency textures for reliable downstream detection, an intuitive solution is to apply super-resolution (SR)~\cite{zhou2023srformer} after weather removal. Unlike general image restoration methods, which primarily focus on noise removal or texture enhancement without explicitly upsampling images, SR methods directly upscale image resolution, thus effectively enlarging small targets to reduce hallucination, as shown in Fig.~\ref{fig:teaser_small}. However, simply cascading weather removal and SR methods fails to address their inherent conflicts: weather removal aims to suppress high-frequency noise, whereas SR attempts to infer high-frequency details from existing textures. Errors introduced in weather removal propagate through SR, resulting in amplified artifacts and inconsistent texture reconstruction. Similarly, fine-tuning deraining models \cite{sun2024restoring} on paired low-resolution (LR) rainy and high-resolution (HR) clean images faces similar issues, as these models inherently struggle to balance high-frequency noise removal and texture recovery. Consequently, there remains a critical need for a method that can simultaneously achieve effective weather noise suppression and accurate SR texture restoration. \emph{In this paper, we focus on image deraining as a representative example to address this challenging scenario.}

\begin{figure*}[ht]
	\hspace{2.5mm}
	\begin{overpic}[width=0.98\linewidth]{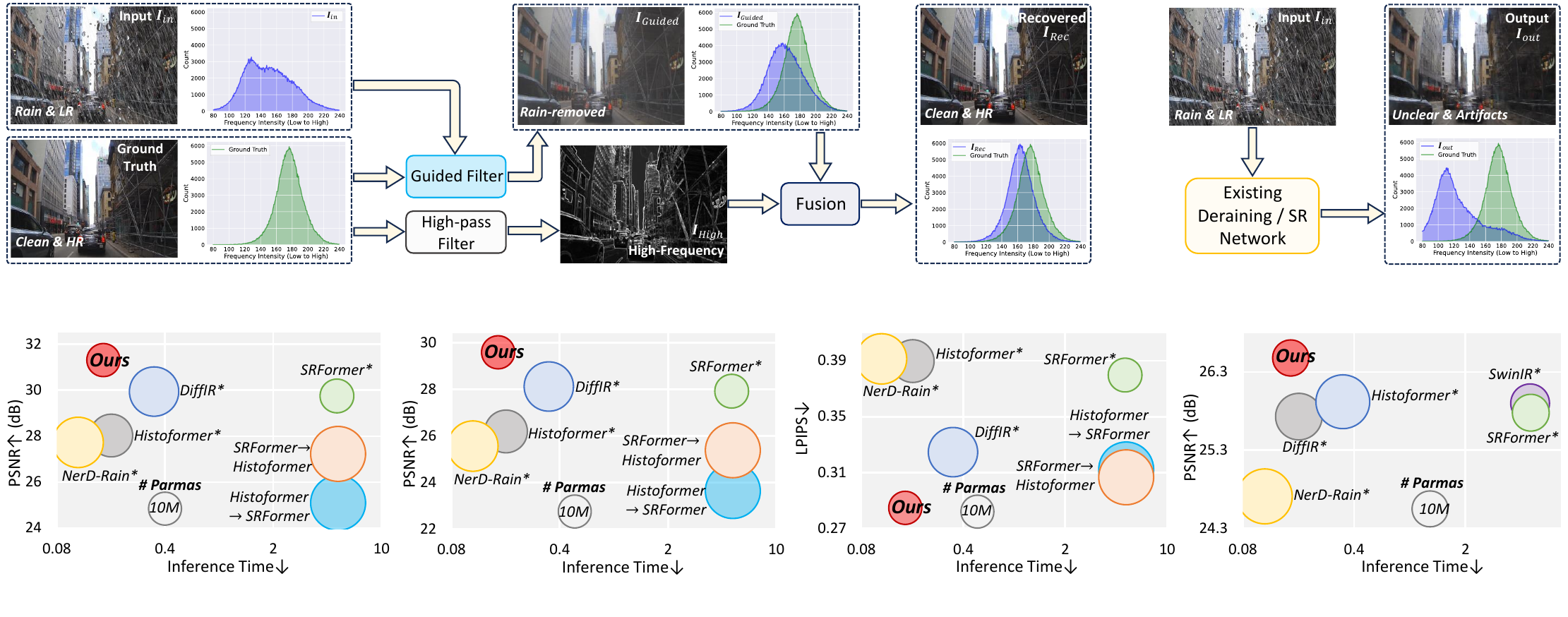}
		
    \put(5.0,0.5){\color{black}{\fontsize{7.8pt}{1pt}\selectfont (a) Synthetic Rain Test set}}
    \put(25.7,0.5){\color{black}{\fontsize{7.8pt}{1pt}\selectfont (b) Synthetic Rain+Raindrop Test set}}
    \put(55.5,0.5){\color{black}{\fontsize{7.8pt}{1pt}\selectfont (c) Real Raindrop Test set}}
    \put(82.5,0.5){\color{black}{\fontsize{7.8pt}{1pt}\selectfont (d) Real Rain Test set}}

    \put(0.0,19.99){\color{black}{\fontsize{7.8pt}{1pt}\selectfont (a) Our insight: guided and high-pass filter for high-frequency textures preserving while removing rain effects.}}
    \put(76.5,19.99){\color{black}{\fontsize{7.8pt}{1pt}\selectfont (b) Pipeline of existing Methods.}}
    
    \put(-2.0,24.1){\color{black}{\fontsize{7.8pt}{1pt}\selectfont {\rotatebox{90}{\textbf{Frequency Analysis}}}}}
    \put(-2.0,4.8){\color{black}{\fontsize{7.8pt}{1pt}\selectfont {\rotatebox{90}{\textbf{Efficiency Trad-off}}}}}

	\end{overpic}
	\caption{(\textbf{Top}) Given a clean and HR image, guided filters can remove high-frequency noises while preserving most high-frequency textures, and high-pass filters further enhance blurred high-frequency edges. (\textbf{Bottom}) Our method achieves the best performance while requiring less cost in terms of speed and model size. (* denotes results of fine-tuning on our training dataset.) }
	\label{fig:teaser2}
\end{figure*}
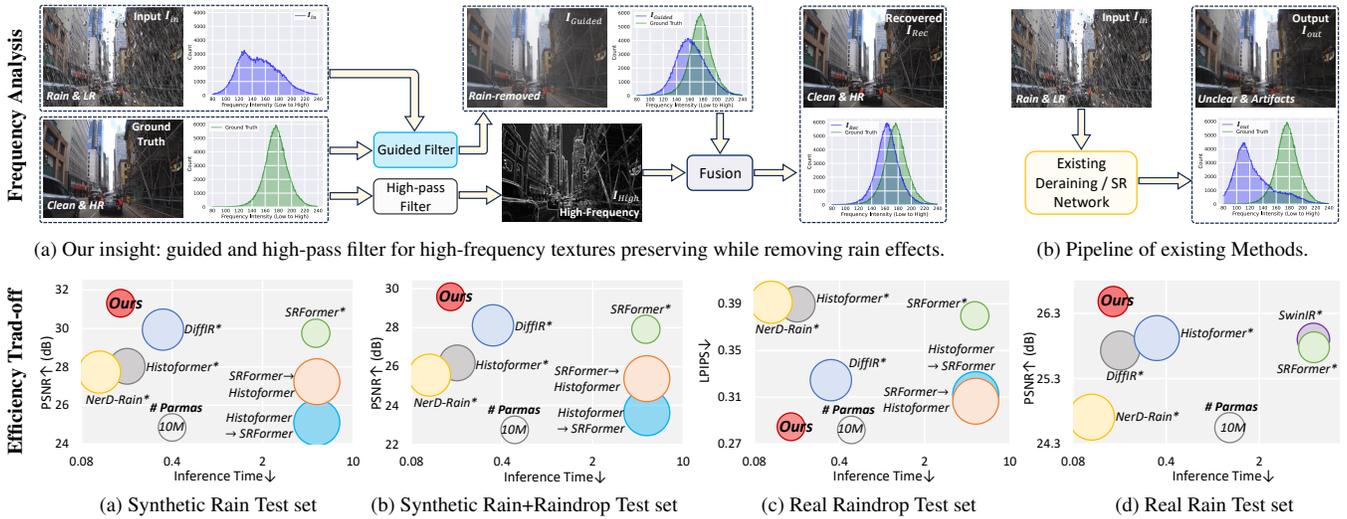

Inspired by guided filters~\cite{he2012guided} and high-pass filters~\cite{khan2016importance}, we try to utilize these to reconstruct rainy LR images. Specifically, as shown in Fig.~\ref{fig:teaser2}, with the help of clean and HR images, we observe guided filter~\cite{he2012guided} can smooth out messy high-frequency noise (\eg, rain and raindrop) while preserving image edge (\eg, high-frequency textures of LR images), thereby aligning the frequency distribution is close to the ground truth (GT). However, high-frequency texture details from restored images remain missing, as shown in the spatial results and frequency distribution. Therefore, a high-pass filter can be further applied to clean HR images to compensate for high-frequency texture features, leading to rain-free and HR outputs. Compared to existing deraining and SR methods, this strategy achieves cleaner and clearer results. Therefore, combining priors with guided and high-pass filters can be a promising solution for SR in rainy weather to address the balance in high-frequency reconstruction.

However, clean and HR images are required by guided and high-pass filters. To achieve this chicken-egg problem, we try to learn content priors close to GT distributions. Specifically, we propose a Diffusion-based High-frequency Guided Model (DHGM) with two phases. In the first phase, we employ encoders to compress contents reflecting true distributions into latent spaces as priors. To leverage latent priors for rain removal and texture reconstruction, we propose a Media Remover (MR) based on guided filters and a Texture Compensator (TC) based on high-pass filters. Recognizing the potential of diffusion models~\cite{ho2020denoising} (DM) to achieve high-quality mappings from randomly sampled Gaussian noise to latent distributions~\cite{rombach2022high,xia2023diffir}, we proceed to the second phase by freezing encoder weights and using DM to learn content distributions within pre-trained latent priors. Simultaneously, we fine-tune our MR and TC from the first phase, training them alongside DM to reconstruct images jointly. As observed at the bottom of Fig.~\ref{fig:teaser_small} and Fig.~\ref{fig:teaser2}, our method can reconstruct clean and HR images from rainy LR images while maintaining efficiency on speed and model size, further improving the accuracy of downstream tasks.

To summarize, our contributions are as follows:
\begin{itemize}
\item We focus on deraining as an example, propose DHGM that explores the challenge of recovering clean and HR images from potential LR images captured in rainy conditions;
\item We propose MR and TC modules based on guided filter and high-pass filter, which direct pre-trained diffusion priors to remove rain-induced noise and recover textures;
\item Experiments on extensive datasets show our method can recover clean and HR images for downstream tasks while requiring less computational cost than existing methods.
\end{itemize}

\section{Related Work}

\begin{figure*}[ht]
	\begin{overpic}[width=0.99\linewidth]{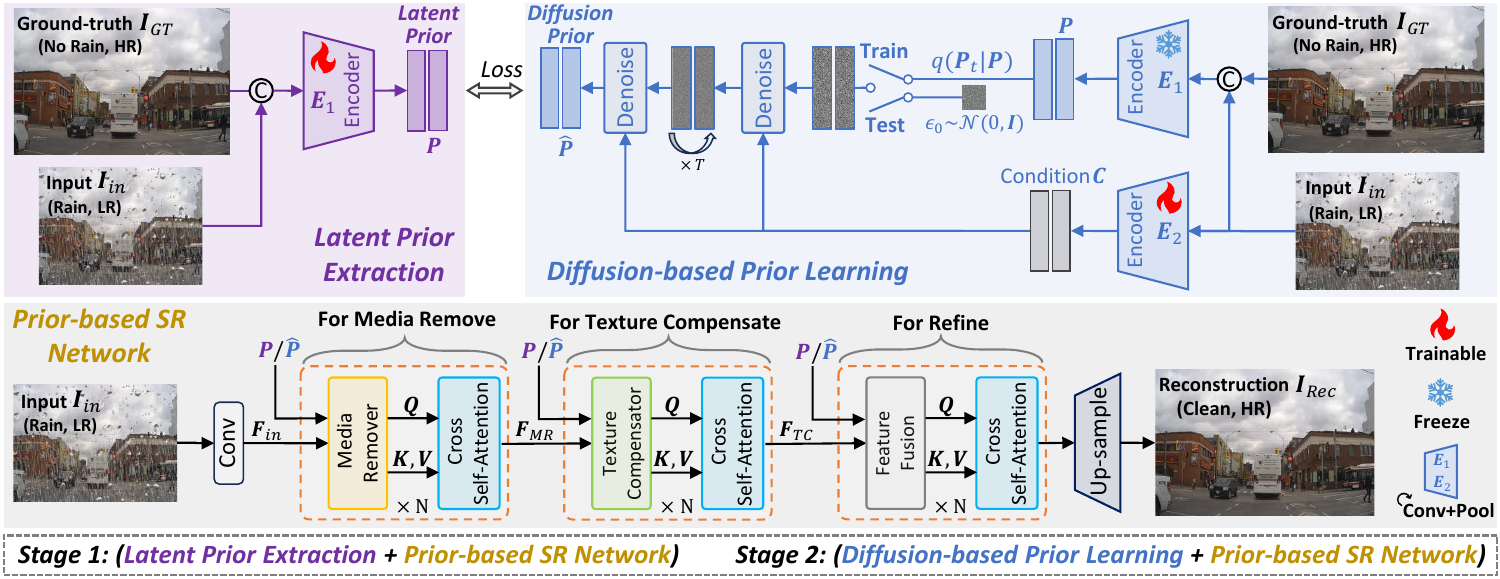}
	\end{overpic}
	\caption{Overview of our method, which utilizes our Media Remover (MR) and Texture Compensator (TC) to guide learned diffusion priors in latent spaces to complete high-frequency rain-induced media removal and high-frequency texture reconstruction.}
	\label{fig: main}
\end{figure*}

\subsection{Restoration in Rainy Weather}
\paragraph{Image Deraining.} To handle rainy images that obstruct the view and are not conducive to downstream tasks~\cite{peng2024lightweight}, a series of specially designed networks~\cite{peng2025boosting} employ strategies like multi-branch~\cite{jiang2020multi}, multi-scale~\cite{chen2023sparse}, and multi-stage~\cite{wang2020rethinking} to achieve end-to-end deraining. Uddin \etal~\cite{uddin2022real} focuses on SR and adjusts rainy tones rather than joint deraining and SR. Subsequently, to enhance the global representation of models, IDT~\cite{xiao2022image} proposes a window-based Transformer, while NeRD-Rain~\cite{chen2024bidirectional} combines multi-scale implicit neural representations. Unlike previous studies, we focus on joint deraining and SR that may occur in rainy weather.

\paragraph{All-in-One Weather Restoration.} Recent attempts have been made to unify complex weather recovery efforts into one network. The all-in-one restoration network~\cite{li2020all} is the first try with multiple task-specific encoders and a shared decoder. TransWeather~\cite{valanarasu2022transweather} improves performance in various rainy conditions through Transformer-based encoder-decoders. WeatherStream~\cite{zhang2023weatherstream}, DTMWR~\cite{patil2023multi}, and WGWS-Net~\cite{zhu2023learning} improve existing models by learning weather-specific degradation data. WeatherDiff~\cite{ozdenizci2023restoring} proposes patch-based denoising diffusion models~\cite{ho2020denoising} to achieve size-agnostic restoration. OneRestore~\cite{guo2024onerestore} proposes a versatile imaging model to simulate possible weather degradation in the environment. Additionally, recent studies also address real data scarcity, fidelity, model lightweight, or modal fusion through techniques such as adaptive filters~\cite{park2023all}, codebooks~\cite{ye2023adverse}, knowledge distillation~\cite{chen2022learning}, and pre-trained language models~\cite{tan2024exploring}. However, texture loss and edge distortion caused by direct weather restoration may seriously affect the contours of small targets at long distances, which is detrimental to downstream tasks, especially including small objects.

\subsection{Image Super-resolution}
Numerous SR algorithms~\cite{li2023survey} have been developed to improve image resolution. Specifically, SRCNN~\cite{dong2015image} first uses a 3-layer convolutional neural network for SR. RCAN~\cite{zhang2018image}, SAN~\cite{dai2019second}, and NLSA~\cite{mei2021image} enhance SR performance by introducing attention mechanisms. WDSR~\cite{yu2018wide} and FDIWN~\cite{gao2022feature} reduce the feature loss of SR processes caused by activation functions. With the development of ViT~\cite{yuan2021tokens}, IPT~\cite{chen2021pre} tries to improve SR by utilizing the Transformer. Then, SwinIR~\cite{liang2021swinir}, FIWHN~\cite{li2024efficient}, and SRFormer~\cite{zhou2023srformer} utilize the Windows-based strategy to reduce the enormous costs caused by the Transformer. OmniSR~\cite{wang2023omni}, ATD~\cite{zhang2024transcending}, and DMNet~\cite{li2025dual} expand the receptive fields of self-attention in the Transformer to improve SR further. These methods promote the advancement of SR, but they default to images without the interference of weathers.

\section{Methods}
As shown in Fig.~\ref{fig: main}, our method with two training phases consists of three components: latent Prior Extraction, Diffusion-based Prior Learning, and Prior-based SR Network. In the first phase, we jointly train the Latent Prior Extraction module for extracting latent priors \hbox{$\boldsymbol{P}$} and the Prior-based SR Network for utilizing \hbox{$\boldsymbol{P}$}. In the second phase, we train the Diffusion-based Prior Learning module for learning diffusion priors \hbox{$\hat{\boldsymbol{P}}$} from \hbox{$\boldsymbol{P}$} while fine-tuning the pre-trained Prior-based SR Network for improved reconstruction. After that, rainy LR inputs \hbox{$\boldsymbol{I}_{in}$~$\in$~$\mathbb{R}^{H \times W\times 3}$} can be recovered to clean and HR outputs \hbox{$\boldsymbol{I}_{Rec}$~$\in$~$\mathbb{R}^{sH \times sW\times 3}$}, $s$ is a scale factor.

\begin{figure*}[ht]
	\begin{overpic}[width=0.99\linewidth]{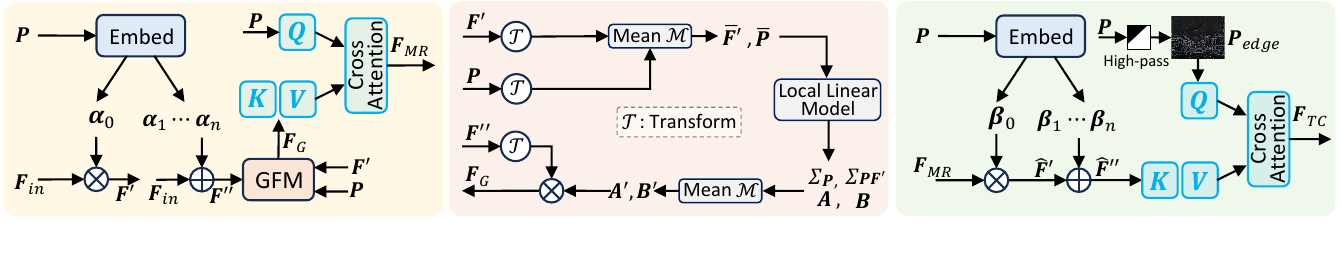}
		\put(8.3,0.3){\color{black}{\fontsize{8pt}{1pt}\selectfont (a) Media Remover (MR)}}
		\put(40.1,0.3){\color{black}{\fontsize{8pt}{1pt}\selectfont (b) Guided Filter Module (GFM)}}
		\put(73.3,0.3){\color{black}{\fontsize{8pt}{1pt}\selectfont (c) Texture Compensator (TC)}}
	\end{overpic}
	\caption{Detailed structure of (a) Media Remover, (b) Guided Filter Module, and (c) Texture Compensator.}
	\label{fig: modules}
\end{figure*}

\subsection{Pre-training Priors Learning (Stage I)}
In the first stage of training, as shown in Fig.~\ref{fig: main}, we focus on jointly training the latent prior extraction module and prior-based SR Network to obtain a compact representation \hbox{$\boldsymbol{P}$} from a hybrid of inputs and ground truth, serving as latent priors. Mixing inputs with ground truth mitigates the distribution gap, preventing excessive divergence that could degrade network performance. For inputs \hbox{${\boldsymbol{I}_{in}}$}, our components remove rain media, reconstruct edge textures, and refine outputs to get final results \hbox{${\boldsymbol{I}_{Rec}}$} via up-sampling.

\paragraph{Media Remover (MR).} As shown in Fig.~\ref{fig: modules} (a), we incorporate our GFM into MR for rain-induced media removal and incorporate cross-attention~\cite{rombach2022high} for feature fusion. Specifically, for input priors \hbox{$\boldsymbol{P}$~$\in$~$\mathbb{R}^{4C}$} and input rainy LR features \hbox{$\boldsymbol{F}_{in}$~$\in$~$\mathbb{R}^{H \times W\times C}$}, we first embed \hbox{$\boldsymbol{P}$} to obtain a set of vectors \hbox{$\left\{ {{\boldsymbol{\alpha}  _0}, {\boldsymbol{\alpha} _1} \ldots {\boldsymbol{\alpha} _n}} \right\}$~$\in$~$\mathbb{R}^{C \times 1\times 1 }$}. To obtain different interaction patterns of priors with \hbox{$\boldsymbol{F}_{in}$}, we fuse prior vectors with features using multiplication and addition, respectively. After that, we obtain two coarse fusion features \hbox{${\boldsymbol{F}', \boldsymbol{F}''}$~$\in$~$\mathbb{R}^{H \times W\times C}$} with different patterns. Then, \hbox{${\boldsymbol{F}', \boldsymbol{F}''}$} are fed into our GFM $\mathcal{G}$ together with priors \hbox{$\boldsymbol{P}$} to obtain outputs \hbox{${\boldsymbol{F}_{G}}$~$\in$~$\mathbb{R}^{H \times W\times C}$} without rain-induced media:
\begin{equation}
	{\boldsymbol{F}'},{\boldsymbol{F}{''}} =  {\boldsymbol{\alpha}_0 \times {\boldsymbol{F}_{in}}},\left\{ {{\boldsymbol{\alpha} _1} \ldots {\boldsymbol{\alpha} _n}} \right\} + {\boldsymbol{F}_{in}},
\end{equation}
\begin{equation}
	{\boldsymbol{F}_G} = \mathcal{G}\left( {{\boldsymbol{F}'},{\boldsymbol{F}{''}},\boldsymbol{P}} \right).
\end{equation}
Next, inspired by the performance of cross-attention~\cite{rombach2022high,li2023cross} for feature fusion, we introduce a cross-attention to fuse \hbox{$\boldsymbol{P}$} and \hbox{${\boldsymbol{F}_{G}}$} to find feature similarity within both. Specifically, we first embed \hbox{${\boldsymbol{F}_{G}}$} and \hbox{$\boldsymbol{P}$} to project into vectors \hbox{$\left\{ {\boldsymbol{Q},\boldsymbol{K},\boldsymbol{V}} \right\}$~$\in$~$\mathbb{R}^{HW\times C}$}:
\begin{equation}
	\boldsymbol{Q},\boldsymbol{K},\boldsymbol{V} = {\mathcal{W}_Q}{\boldsymbol{F}_G} \times \boldsymbol{P},{\mathcal{W}_K}{\boldsymbol{F}_G},{\mathcal{W}_V}{\boldsymbol{F}_G},
\end{equation}
where \hbox{${\mathcal{W}_Q}$}, \hbox{${\mathcal{W}_K}$}, and \hbox{${\mathcal{W}_V}$} are convolution operations. Next, we utilize cross-attention to achieve information fusion and explore the relationship between \hbox{$\boldsymbol{P}$} and \hbox{${\boldsymbol{F}_{G}}$}:
\begin{equation}
	{\rm{CrossAttention}}({\boldsymbol{Q}},{\boldsymbol{K}},{\boldsymbol{V}}) \!=\! {\boldsymbol{V}}{\rm{Softmax}}({\boldsymbol{Q}}{{\boldsymbol{K}}^T}/\gamma  ),
\end{equation}
where $\gamma$ is a learnable factor. Finally, we reconstruct features \hbox{${\boldsymbol{F}_{MR}}$} without rain-induced media using cross-attention. Details of GFM responsible for guiding prior removal of rain-related media are described in the following paragraph.

\paragraph{Guided Filter Module (GFM).} Inspired by the concept of guided filter~\cite{he2012guided}, we propose a GFM to bootstrap priors for removing rain-induced media while preserving structural details. As shown in Fig.~\ref{fig: modules} (b), the process begins with two coarse features \hbox{$\boldsymbol{F}'$} and \hbox{$\boldsymbol{F}''$}, modulated by priors \hbox{$\boldsymbol{P}$}. First, a transformation function $\mathcal{T}$ is applied to \hbox{$\boldsymbol{F}'$} and \hbox{$\boldsymbol{P}$}, followed by a mean filtering $\mathcal{M}$ with a radius $r$, producing the smoothed representations $\overline {\boldsymbol{F}'}$ and $\overline {\boldsymbol{P}}$:
\begin{equation} 
\overline {\boldsymbol{F}'} , \overline {\boldsymbol{P}} = \mathcal{M}\left( \mathcal{T}\left(\boldsymbol{F}'\right), r \right), \mathcal{M}\left( \mathcal{T}\left(\boldsymbol{P}\right), r \right). \end{equation}
This step captures local dependencies between features and priors, enabling the network to model interactions across different spatial scales. Next, we refine the interaction between input features \hbox{$\boldsymbol{F}'$} and priors \hbox{$\boldsymbol{P}$} by computing their filtered correlation terms, ${\boldsymbol{PF}'}$ and ${\boldsymbol{P}^2}$, which help to characterize the relationship between features and priors, providing an accurate representation of low-frequency structures: 
\begin{equation} 
\overline {\boldsymbol{PF}'} , \overline {\boldsymbol{P}^2} = \mathcal{M}\left( \boldsymbol{F}' \cdot \boldsymbol{P}, r \right), \mathcal{M}\left( \boldsymbol{P} \cdot \boldsymbol{P}, r \right).
\end{equation}
We then calculate coefficients \hbox{$\boldsymbol{A}$} and \hbox{$\boldsymbol{B}$}, which quantify priors' influence on inputs and low-frequency details, respectively, helping accurately extract background details: 
\begin{equation}
\sum \boldsymbol{P}, \sum \boldsymbol{P} \boldsymbol{F}' = \overline {\boldsymbol{P}^2} - \overline {\boldsymbol{P}} \cdot \overline {\boldsymbol{P}}, \overline {\boldsymbol{PF}'} - \overline {\boldsymbol{P}} \cdot \overline {\boldsymbol{F}'} , \end{equation} 
\begin{equation} \hbox{$\boldsymbol{A}$} = \frac{\sum \boldsymbol{P} \boldsymbol{F}'}{\left( \sum \boldsymbol{P} + \epsilon \right)}, \hbox{$\boldsymbol{B}$} = \overline {\boldsymbol{F}'} - \hbox{$\boldsymbol{A}$} \cdot \overline {\boldsymbol{P}}. 
\end{equation}
To ensure consistency in smoothing, we apply mean filtering, yielding learned guidance coefficients \hbox{$\overline {\hbox{$\boldsymbol{A}$}}$} and \hbox{$\overline {\hbox{$\boldsymbol{B}$}}$}. They are then fused with \hbox{$\boldsymbol{F}''$} through element-wise modulation to generate guided features \hbox{$\boldsymbol{F}_G$}, which is free from rain:
\begin{equation} 
\overline {\hbox{$\boldsymbol{A}$}} = \mathcal{M}\left( \hbox{$\boldsymbol{A}$} \right), \overline {\hbox{$\boldsymbol{B}$}} = \mathcal{M}\left( \hbox{$\boldsymbol{B}$} \right), 
\end{equation} 
\begin{equation} {\boldsymbol{F}_G} = \overline {\hbox{$\boldsymbol{A}$}} \cdot \boldsymbol{F}'' + \overline {\hbox{$\boldsymbol{B}$}}.
\end{equation}
Through the guide of $\boldsymbol{\overline A}$, which reflects priors' influence, and $\boldsymbol{\overline B}$, which reflects background details, GFM removes rain-induced media while preserving critical structural details.

\begin{table*}[t!]
	\setlength\tabcolsep{2pt}
	\centering
    \resizebox{1\textwidth}{!}{
    \begin{tabular}{l|c|c|ccc|ccc|ccc}
    \hline
    \toprule 
    \rowcolor{lightgray}
    & & & \multicolumn{3}{c|}{RainDS-Syn-Rain} 
    & \multicolumn{3}{c|}{RainDS-Syn-RD-Rain} 
    & \multicolumn{3}{c}{Raindrop} 
    \\ 
    \cmidrule{4-12}
        \rowcolor{lightgray}
        \multicolumn{1}{l|}{\multirow{-2}{*}{Methods}} 
        & \multicolumn{1}{c|}{\multirow{-2}{*}{Params}}
        & \multicolumn{1}{c|}{\multirow{-2}{*}{Speed}}
        & PSNR/SSIM    & LPIPS$\downarrow$  & DISTS$\downarrow$ 
        
        & PSNR/SSIM    & LPIPS$\downarrow$  & DISTS$\downarrow$ 
        & PSNR/SSIM    & LPIPS$\downarrow$  & DISTS$\downarrow$ 
        \\ 
        \hline\hline
        Bicubic (LR+Rainy)    & -   & -   & 22.25/0.6525  & 0.4210   & 0.2513 & 19.41/0.5571 & 0.4892  & 0.2698 & 22.69/0.7027  & 0.3674  & 0.2002 \\
        Histoformer$\rightarrow$SwinIR    & 28.5M   & 5.29s   & 25.42/0.7404  & 0.4305   & 0.2523 & 23.62/0.6804 & 0.4749  & 0.2829 & 24.52/0.7355 & 0.3130  & 0.1424 \\
        Histoformer$\rightarrow$SRFormer    & 27.1M  & 5.24s   & 25.10/0.7210  & 0.4311   & 0.2525  & 23.63/0.6804 & 0.4753 & 0.2830 & 24.53/0.7364 & 0.3122  & 0.1422\\
        SwinIR$\rightarrow$Histoformer   & 28.5M  & 5.29s  & 27.20/0.7998  & 0.3195   & 0.1783 & 25.35/0.7553 & 0.3762  & 0.2125 & 24.40/0.7359 & 0.3066  & \underline{0.1385}\\
        SRFormer$\rightarrow$Histoformer   & 27.1M   & 5.24s  & 27.23/0.8005  & 0.3190   & 0.1786 & 25.38/0.7563 & 0.3758  & 0.2122  & 24.45/0.7374 & \underline{0.3047}  & 0.1387\\
    
        Fine-tuned Histoformer  & 16.6M  & 0.18s & 28.03/0.8274  & 0.2818   & 0.1584 & 26.19/0.7883 & 0.3373  & 0.1913 & 24.26/0.7064 & 0.3898  & 0.1982\\
    
        Fine-tuned NeRD-Rain  & 22.9M & \textbf{0.11s}  & 27.73/0.8206  & 0.2962   & 0.1642 & 25.56/0.7705 & 0.3642 & 0.1945 & 22.66/0.6825 & 0.3914  & 0.2016\\
    
    
        Fine-tuned SRFormer   & \underline{10.5M}    & 5.15s & 29.75/0.8751  & 0.2658   & 0.1031 & 27.92/0.8444 & 0.2465  & \underline{0.1222}   & 24.45/0.7314 & 0.3798  & 0.1905\\
    
        Fine-tuned DiffIR   & 22.0M   & 0.34s & \underline{29.94}/\underline{0.8795}  & \underline{0.1950}   & \underline{0.1000} & \underline{28.12}/\underline{0.8517} & \underline{0.2375}  & 0.1225  & \underline{25.42}/\underline{0.7401} & 0.3246  & 0.1597\\
    
        \hline\hline
        \textbf{Ours}    & \textbf{10.1M}   & \underline{0.16s} & \textbf{31.28/0.9067}  & \textbf{0.1575}   & \textbf{0.0787} & \textbf{29.63/0.8863} & \textbf{0.1916}  & \textbf{0.0959}  & \textbf{26.21/0.7709} & \textbf{0.2862}  & \textbf{0.1378}\\
        \toprule
    \end{tabular}}
    \caption{Quantitative results of ours with sequentially performed deraining and SR methods, fine-tuned single deraining, all-in-one weather restoration, SR, and general restoration methods on deraining, deraining \& raindrop removal, and raindrop removal test sets at scale of $\times 2$ (with the resolution of 720$\times$480). Best and second-best results are emphasized in \textbf{bold} and \underline{underlined}.}
    \label{tab: synthetic_performance}
    \end{table*}

\paragraph{Texture Compensator (TC).} As shown in Fig.\ref{fig: modules} (c), we propose a TC module to refine high-frequency parts of features after rain media removal. Inputs consist of features \hbox{$\boldsymbol{F}_{MR}$$\in$$\mathbb{R}^{H \times W \times C}$} from the MR module, which has undergone rain-related media removal, along with priors \hbox{$\boldsymbol{P}$$\in$$\mathbb{R}^{4C}$}. We begin by embedding prior \hbox{$\boldsymbol{P}$} into a set of sub-priors \hbox{$\left\{ {{\boldsymbol{\beta} _0}, {\boldsymbol{\beta} _1} \ldots {\boldsymbol{\beta} _n}} \right\}$~$\in$~$\mathbb{R}^{C \times 1\times 1 }$}. These sub-priors are then fused with \hbox{$\boldsymbol{F}_{MR}$} to obtain coarse features ${\hat{\boldsymbol{F}}{''}}$: 
\begin{equation}
	{\hat{\boldsymbol{F}}{''}} = {\boldsymbol{\beta}_0 \times {\boldsymbol{F}_{MR}}} + \left\{ {{\boldsymbol{\beta} _1} \ldots {\boldsymbol{\beta} _n}} \right\} + {\boldsymbol{F}_{MR}}.
\end{equation}
To inject missed textures, we decide to apply a high-pass filter for extracting high-frequency parts from priors \hbox{$\boldsymbol{P}$}. This is achieved by performing a 1D Discrete Cosine~\cite{khayam2003discrete} Transform (DCT) along channels, transforming \hbox{$\boldsymbol{P}$} from the spatial domain to the frequency domain. After this, we perform an inverse transform (IDCT) to recover spatial domain features, thus extracting edge features \hbox{${\boldsymbol{P}_{edge}}$}: 
\begin{equation}
	{\boldsymbol{X}_k} \!=\! \mathcal{DCT}\left( \boldsymbol{P} \right) \!=\! \sum\limits_{n = 0}^{N - 1} {{\boldsymbol{P}_n}} \cdot \cos \left( {\frac{\pi }{N}\left( {n \!+\! \frac{1}{2}} \right)k} \right),
\end{equation}
\begin{equation}
	{\boldsymbol{X}_{h}} = 
	\begin{cases} 
		0, &  k < k_{cutoff} \\
		{{\boldsymbol{X}_k}}, &  k \geq k_{cutoff},
	\end{cases}
\end{equation}
\begin{equation}
	{\boldsymbol{P}_{edge}} \!\!=\! \mathcal{IDCT}\!\left( \boldsymbol{X}_{h} \right) \!=\!\! \sum\limits_{k = 0}^{N \!-\! 1}\! \boldsymbol{X}_{h}^k \cdot \cos \left( {\!\frac{\pi }{N}\!\left(\! {n \!+\! \frac{1}{2}} \right)\!k} \!\right),
\end{equation}
where \hbox{${\boldsymbol{P}_n}$} is $n$-th element of \hbox{$\boldsymbol{P}$}, $N$ is channel counts, $k = 0,1, \ldots ,N - 1$, \hbox{${\boldsymbol{X}_k}$} represents DCT coefficients, \hbox{${\boldsymbol{X}_h}$} represents high-frequency components of \hbox{$\boldsymbol{P}$}, and 
$k_{cutoff}$ is the cut-off frequency. To get matched dimensions, we reshape \hbox{${\boldsymbol{P}_{edge}}$} to \hbox{$\mathbb{R}^{1 \times 1 \times C}$} and feed it with coarse features \hbox{${\hat{\boldsymbol{F}}{''}}$} into a cross-attention to further explore high-frequency components and enhance edges~\cite{li2024efficientface}. This enables the reconstruction of sharp edges in outputs: 
\begin{equation} \boldsymbol{Q}, \boldsymbol{K}, \boldsymbol{V} = {\mathcal{W}_Q}{\hat{\boldsymbol{F}}{''}} \times {\boldsymbol{P}_{edge}},{\mathcal{W}_K}{\hat{\boldsymbol{F}}{''}},{\mathcal{W}_V}{\hat{\boldsymbol{F}}{''}}, 
\end{equation} 
\begin{equation} 
{\rm{CrossAttention}}({\boldsymbol{Q}}, \!{\boldsymbol{K}}, \!{\boldsymbol{V}}) \!=\! {\boldsymbol{V}}{\rm{Softmax}}({\boldsymbol{Q}}{{\boldsymbol{K}}^T}\!/\gamma). \end{equation}
Through these operators, we compensate for edges missed in inputs. Next, as shown in Fig.\ref{fig: main}, we use product to fuse \hbox{${\boldsymbol{F}_{TC}}$} with priors \hbox{$\boldsymbol{P}$}, followed by cross-attention for feature refinement. Final results, \hbox{$\boldsymbol{I}_{rec}$} are obtained through up-sampling, producing texture-compensated and rain-free images.


\subsection{Prior-based Guided Restoration (Stage II)}
In the second stage, as shown in Fig.~\ref{fig: main}, we freeze pre-trained encoder \hbox{${\boldsymbol{E}_{1}}$} weights and output priors \hbox{$\hat{\boldsymbol{P}}$} extracted from GT. We hope to accurately estimate \hbox{$\hat{\boldsymbol{P}}$} close to \hbox{$\boldsymbol{P}$} without GT for fine-tuning our prior-based SR network. Inspired by the ability of DDIM~\cite{ho2020denoising} to generate high-quality images from random noises, surpassing StyleGAN~\cite{karras2019style} and VQGAN~\cite{esser2021taming}, we adopt it for prior learning. Furthermore, instead of conventional DDIM, we follow the latent-space diffusion denoising approach~\cite{xia2023diffir}, reducing result randomness and avoiding the inefficiency of pixel-wise generation~\cite{li2025self,pengtowards}.

\paragraph{Diffusion and Denoising Process.} As shown in Fig.~\ref{fig: main}, for the diffusion process, we obtain latent prior \hbox{$\boldsymbol{P}$} from froze encoder \hbox{${\boldsymbol{E}_{1}}$}, and progressively add Gaussian noise on \hbox{$\boldsymbol{P}$}:
\begin{equation}
	q\left( {{\boldsymbol{P}_t}|\boldsymbol{P}} \right) = \mathcal{N}\left( {{\boldsymbol{P}_t};\sqrt {{{\overline \alpha  }_t}} \boldsymbol{P},\left( {1 - {{\overline \alpha  }_t}} \right)\boldsymbol{I}} \right),
\end{equation}
where ${\boldsymbol{P}_t}$ is a noised prior at time-step $t$, $\mathcal{N}$ is a Gaussian distribution, ${\alpha _t} = 1 - {\beta _t}$, ${\overline \alpha  _t} = \prod\nolimits_{i = 0}^t {{\alpha _t}} $, ${\beta _t}$ is a scale factor to control the variance of noises, and $\boldsymbol{I}$ is an identity matrix. For the denoising process, following the Markov chain, the reverse process from ${\boldsymbol{P}_t}$ to ${\boldsymbol{P}_{t-1}}$ can be formulated as:
\begin{equation}
	p\left( {{\boldsymbol{P}_{t - 1}}|{\boldsymbol{P}_t},{\boldsymbol{P}_0}} \right) = \mathcal{N}\left( {{\boldsymbol{P}_{t - 1}};{\boldsymbol{\mu} _t}\left( {{\boldsymbol{P}_t},{\boldsymbol{P}_0}} \right),\sigma _t^2\boldsymbol{I}} \right),
\end{equation}
\begin{equation}
	{\boldsymbol{\mu} _t}\!\left(\! {{\boldsymbol{P}_t}, \!{\boldsymbol{P}_0}} \!\right) \!=\!\! \frac{1}{{\sqrt {{\alpha _t}} }}\!\left(\!\! {{\boldsymbol{P}_t} \!-\! \boldsymbol{\epsilon} \frac{{1 \!-\! {\alpha _t}}}{{\sqrt {1 \!-\! {{\overline \alpha  }_t}} }}} \!\right)\!, \!\sigma _t^2 \!=\! \frac{{1 \!\!-\! {{\overline \alpha  }_{t \!-\! 1}}}}{{1 \!\!-\! {{\overline \alpha  }_t}}}{\beta _t},
\end{equation}
where $\boldsymbol{\epsilon}$ is noises in ${\boldsymbol{P}_t}$. In our denoising phase, we use encoder ${\boldsymbol{E}_2}$ to encode input images ${\boldsymbol{I}_{in}}$ to output conditional features ${\boldsymbol{C}}$ to control the range of noise predicted by the denoising network and denoise ${\boldsymbol{P}_t}$ stepwise:
\begin{equation}
	{\boldsymbol{P}_{t \!-\! 1}} \!\!=\! \frac{1}{{\sqrt {{\alpha _t}} }}\left(\!\! {{\boldsymbol{P}_t} \!-\! \frac{{(1 \!-\! {\alpha _t}})\boldsymbol{\epsilon}_\theta }{{\sqrt {1 \!-\! {{\overline \alpha  }_t}} }}\!\!\left( {{\boldsymbol{P}_t},\boldsymbol{C},t} \right)} \!\!\right) \!+\! \sqrt {1 \!\!-\! {\alpha _t}}\boldsymbol{\epsilon}_t  ,
\end{equation}
where $\boldsymbol{\epsilon_t }$ is estimated noise $\boldsymbol{\epsilon}$ of each step, $\boldsymbol{\epsilon_t } \!\sim\! \mathcal{N}\left( {0,\boldsymbol{I}} \right)$. With $T$ iterations of the above sampling, predicted priors ${\hat{\boldsymbol{P}}}$ can be generated to fine-tune our restoration network. 

\paragraph{Inference.} Without access to the ground truth during inference, we follow DDIM~\cite{song2020denoising} and randomly sample Gaussian noise $\boldsymbol{\epsilon_0 } \!\sim\! \mathcal{N}\left( {0,\boldsymbol{I}} \right)$ as noised inputs. The noises, along with the condition 
\hbox{$\boldsymbol{C}$} extracted from ${\boldsymbol{I}_{in}}$, are fed into the denoising process. After $T$ iterations, we get generated 
\hbox{$\hat{\boldsymbol{P}}$}, which serves as guidance for our prior-based SR network, enabling plausible inference without requiring the ground truth.

\subsection{Loss Function}
\paragraph{Stage I.} We jointly train the latent prior extraction module and prior-based SR network. Our training loss \hbox{${\mathcal{L}_{S1}}$} is:
\begin{equation}
	{\mathcal{L}_{S1}} = {{{\left\| {\boldsymbol{I}_{Rec} - \boldsymbol{I}_{GT}} \right\|}_1}},
\end{equation}
where $\boldsymbol{I}_{Rec}$ is recovered image, $\boldsymbol{I}_{GT}$ is the ground truth. 

\paragraph{Stage II.} Following previous works~\cite{rombach2022high,xia2023diffir}, we conduct denoising in latent spaces. Unlike the time-consuming mode of traditional DM, which denoises full images. This strategy allows DM to run denoising iterations to obtain denoising results, which are then sent to prior-based SR networks for joint training. \hbox{${\mathcal{L}_{S2}}$} is:
\begin{equation}
	{\mathcal{L}_{S2}} = {{{\left\| {\boldsymbol{I}_{Rec} - \boldsymbol{I}_{GT}} \right\|}_1}} + {{{\left\| {{\hat{\boldsymbol{P}}} - {\boldsymbol{P}}} \right\|}_1}},
\end{equation}
where \hbox{$\boldsymbol{P}$} is prior extracted by encoder ${{\boldsymbol{E}_1}}$ in the first stage, \hbox{$\hat{\boldsymbol{P}}$} is prior estimated by our diffusion model.

\begin{figure*}[ht]
	\hspace{2mm}
	\begin{overpic}[width=0.98\linewidth]{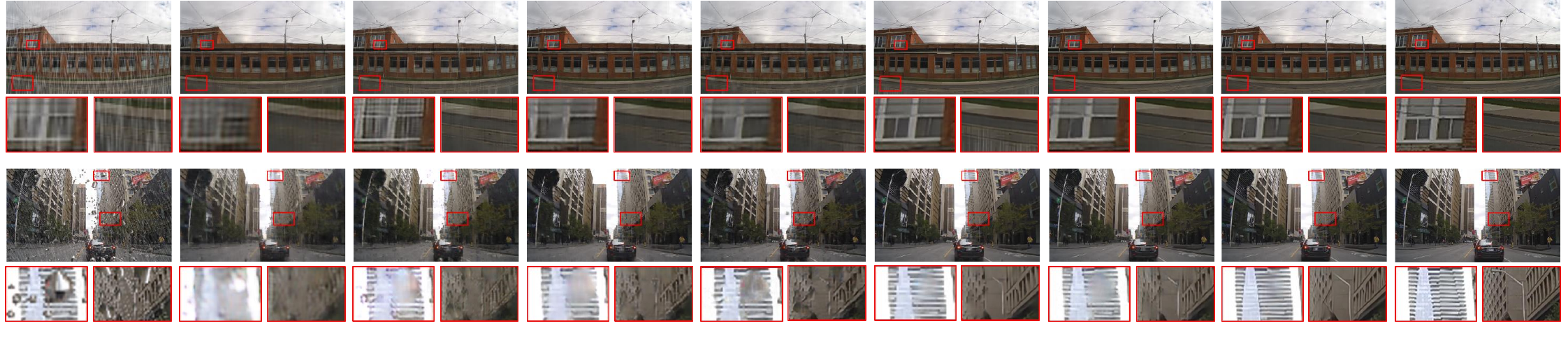}
		\put(2.6,-0.5){\color{black}{\fontsize{7.5pt}{1pt}\selectfont LR+Rainy}}
		\put(14.2,-0.5){\color{black}{\fontsize{7.5pt}{1pt}\selectfont His\scalebox{0.75}{$\rightarrow$}SR}}
		\put(25.3,-0.5){\color{black}{\fontsize{7.5pt}{1pt}\selectfont SR\scalebox{0.75}{$\rightarrow$}His}}
		\put(34.8,-0.5){\color{black}{\fontsize{7.5pt}{1pt}\selectfont Histoformer*}}
		\put(46.4,-0.5){\color{black}{\fontsize{7.5pt}{1pt}\selectfont NeRD-Rain*}}
		\put(57.8,-0.5){\color{black}{\fontsize{7.5pt}{1pt}\selectfont SRformer*}}
		\put(69.6,-0.5){\color{black}{\fontsize{7.5pt}{1pt}\selectfont DiffIR*}}
		\put(81.5,-0.5){\color{black}{\fontsize{7.5pt}{1pt}\selectfont \textbf{Ours}}}
		\put(89.9,-0.5){\color{black}{\fontsize{7.5pt}{1pt}\selectfont Ground Truth}}
		
		\put(-2.0,1.2){\color{black}{\fontsize{7.8pt}{1pt}\selectfont {\rotatebox{90}{\textbf{Rain+Raindrop}}}}}
		\put(-2.0,15.5){\color{black}{\fontsize{7.8pt}{1pt}\selectfont {\rotatebox{90}{\textbf{Rain}}}}}
	\end{overpic}
	\caption{Qualitative comparisons with existing methods on synthesized test sets at the scale of $\times 2$. *: Method after fine-tuning.}
    \label{fig: syn_visual}
\end{figure*}
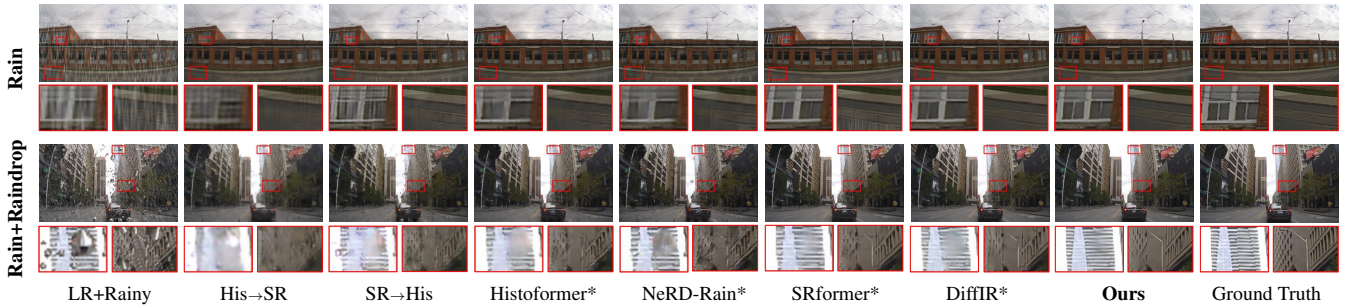

\section{Experiments}
\subsection{Datasets and Evaluation Metrics}
We train our model on datasets containing multiple rainy conditions, which consider raindrops on the camera sensor, heavy rain, and heavy rain with raindrops, respectively. RainDrop~\cite{qian2018attentive} consists of 861 training images and 307 test images with on-camera raindrops. RainDS~\cite{quan2021removing} includes synthetic parts and real parts, which the synthetic part and real part including 3000 training images and 600 test images with rain, on-camera raindrops, and rain with raindrops, and 450 training images and 294 test images with rain, on-camera raindrops, and rain with raindrops, respectively. For evaluation metrics, we calculate PSNR~\cite{wang2002universal} and SSIM~\cite{wang2002universal}, LPIPS~\cite{zhang2018unreasonable}, DISTS~\cite{ding2020image}, and NIQE~\cite{mittal2012making}.


\subsection{Implementation Details}
We implement all experiments in the Pytorch framework with one NVIDIA RTX4090 GPU. During training, we set the batch size to 8, the learning rate to \hbox{$2 \times {10^{ - 4}}$}, and the patch size to \hbox{$64 \times 64$}. We use Adam optimizer with \hbox{${\beta _{\rm{1}}}$}=0.9, \hbox{${\beta _{\rm{2}}}$}=0.99 to train 500k iterations. Our model sets channel counts to 64, \hbox{${N}$} to 12, and time-steps to 4 in our diffusion model. {\emph{Since directly performing deraining and SR on original rainy images is impractical due to the absence of corresponding clean HR ground truth, for experimental validation, we construct paired data rainy LR by downsampling, including both synthetic and real adverse weather.} For fine-tuned SR~\cite{liang2021swinir,zhou2023srformer}, restoration~\cite{xia2023diffir}, deraining~\cite{valanarasu2022transweather,sun2024restoring} methods, we load official pre-trained weights and fine-tune 500k iterations with same hyperparameters in their paper on our training sets. For downstream tasks, we utilize YOLOv8~\cite{varghese2024yolov8} for object detection.

\subsection{Comparison of Deraining under LR Scenes}
We select fine-tuned deraining method (\eg NerD-Rain~\cite{chen2024bidirectional}), all-in-one weather restoration method (\eg Histoformer~\cite{sun2024restoring}), fine-tuned SR methods (\eg SwinIR~\cite{liang2021swinir}, SRFormer~\cite{zhou2023srformer}), and fine-tuned general image restoration methods (\eg DiffIR~\cite{xia2023diffir}). Additionally, we alternate the order of deraining and SR methods for fair comparison. For fine-tuned deraining and restoration methods without up-sampling modules, we first up-sample inputs to match the size of the ground truth before passing them to networks.

\paragraph{Comparison on Rainy Datasets.} As present in Table~\ref{tab: synthetic_performance}, fine-tuned and alternating methods consistently lag behind our method across all metrics, including visual perception and structural metrics. Furthermore, our method show efficiency, with fewer Params and faster inference, especially compared to alternating methods. Besides, as shown in Fig.~\ref{fig: syn_visual}, visual comparisons reveal that existing methods fail to remove rain-induced media or introduce artifacts. In contrast, our method effectively handles this scene, producing high-quality results.

\begin{table}[t!]
	\setlength\tabcolsep{2pt}
	\centering
	\resizebox{0.48\textwidth}{!}{
		\begin{tabular}{l|cc|cc}
			\hline
			\toprule 
			\rowcolor{lightgray}
			& \multicolumn{2}{c|}{RainDS-(RD+Rain)} 
			& \multicolumn{2}{c}{RainDS-(Rain)} 
			\\ 
			\cmidrule{2-5}
			\rowcolor{lightgray}
			\multicolumn{1}{l|}{\multirow{-2}{*}{Methods}} 
			& PSNR/SSIM & NIQE$\downarrow$ 
			
			& PSNR/SSIM & NIQE$\downarrow$ 
			\\ 
			\hline\hline
			Fine-tuned SwinIR    & \underline{22.84}/0.6061  & 7.3883  & 25.90/0.6796   & 6.9523 \\
			
			Fine-tuned SRFormer    & 22.82/0.6040  & 5.1953  & 25.78/0.6758  & \underline{4.9914} \\
			
			Fine-tuned NeRD-Rain    & 20.54/0.5764  & 6.3025  & 24.71/0.6620   & 5.8472  \\
			
			Fine-tuned Histoformer    & 22.77/0.6173  & \underline{5.1382}  & 25.73/0.6923   & 5.7635 \\
			
			Fine-tuned DiffIR  & 22.83/\underline{0.6256}  & 7.3596  & \underline{25.92}/\underline{0.7107}   & 7.0178 \\ 
			
			\hline\hline
			\textbf{Ours}    & \textbf{23.21/0.6522}  & \textbf{4.9628}  & \textbf{26.48/0.7273}   & \textbf{4.9401} \\
			\toprule
	\end{tabular}}
    \caption{Quantitative comparison at a scale of $\times 2$ (with the resolution of 1296$\times$728), where derain \& raindrop removal and derain evaluations are shown on the left and right sides.}
    \label{tab: real_performance}
\end{table}

In Table~\ref{tab: synthetic_performance} and Table~\ref{tab: real_performance}, we conduct experiments on real rainy test sets (LR is synthetic), including deraining, removing raindrops, and deraining \& removing raindrops. Since fine-tuned methods generally outperform alternating methods. In Table~\ref{tab: synthetic_performance}, we only show fine-tuned methods for simplicity. Our method achieves superior results in evaluations. 

\paragraph{Extension to More Weather Conditions.} As shown in Fig.~\ref{fig: visual_expand}, we further validate the ability of our method to handle potential LR scenes in more adverse weather, including snow and rain with haze weather. Visual comparisons show that our method can handle multiple weather LR conditions, and reconstructed images have sharper textures and cleaner backgrounds, with the potential to be extended to more weather.

\subsection{Ablation Study}
We focus on two aspects in ablations: (i) whether priors are needed and how to learn priors. (ii) Are guided filters, high-pass filters, and cross-attention in our Media Remover (MR) and Texture Compensator (TC) effective?

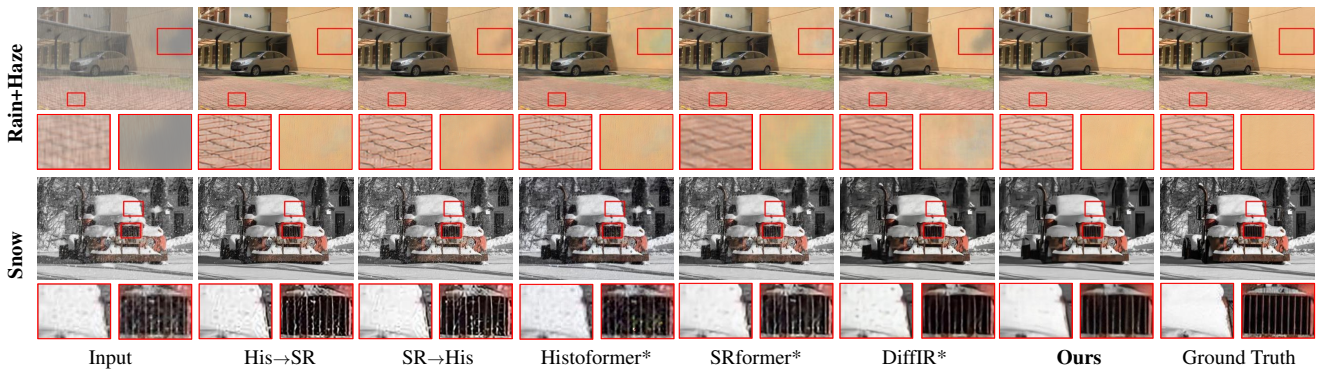
\begin{figure*}[t!]
	\hspace{2mm}
	\begin{overpic}[width=0.96\linewidth]{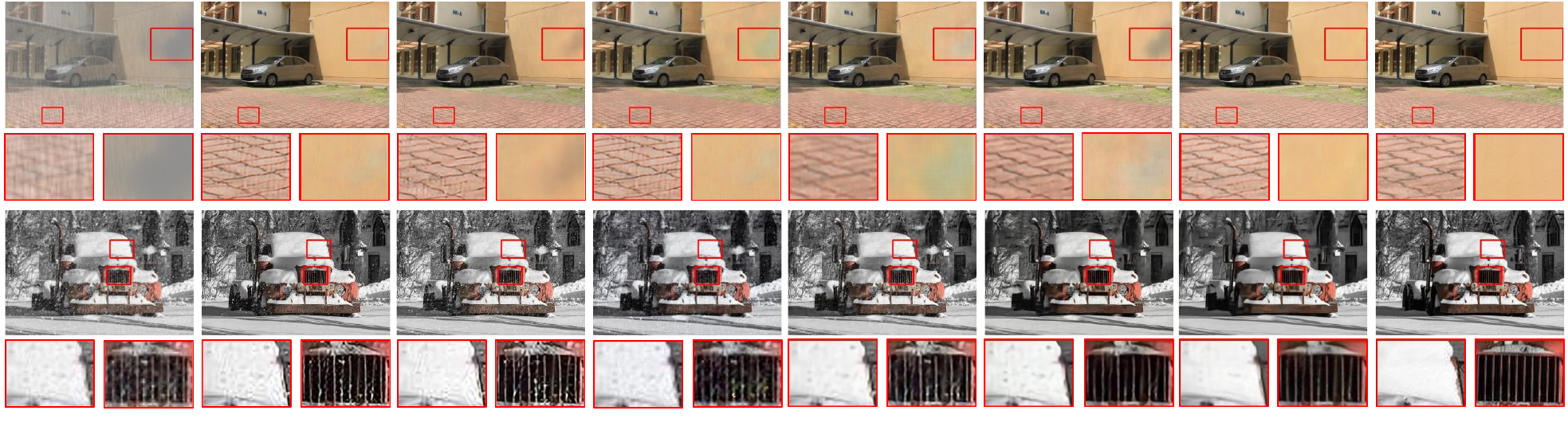}
		\put(4.3,0){\color{black}{\fontsize{7.8pt}{1pt}\selectfont Input}}
		\put(16.4,0){\color{black}{\fontsize{7.8pt}{1pt}\selectfont His\scalebox{0.80}{$\rightarrow$}SR}}
		\put(28.7,0){\color{black}{\fontsize{7.8pt}{1pt}\selectfont SR\scalebox{0.80}{$\rightarrow$}His}}
		\put(39.5,0){\color{black}{\fontsize{7.8pt}{1pt}\selectfont Histoformer*}}
		\put(52.7,0){\color{black}{\fontsize{7.8pt}{1pt}\selectfont SRformer*}}
		\put(66.1,0){\color{black}{\fontsize{7.8pt}{1pt}\selectfont DiffIR*}}
		\put(79.7,0){\color{black}{\fontsize{7.8pt}{1pt}\selectfont \textbf{Ours}}}
		\put(89.5,0){\color{black}{\fontsize{7.8pt}{1pt}\selectfont Ground Truth}}
		
		\put(-2.0,6.6){\color{black}{\fontsize{7.8pt}{1pt}\selectfont {\rotatebox{90}{\textbf{Snow}}}}}
		\put(-2.0,17.2){\color{black}{\fontsize{7.8pt}{1pt}\selectfont {\rotatebox{90}{\textbf{Rain+Haze}}}}}
	\end{overpic}
	\caption{We show our method's capability to resolve possible LR images under more weather, like rain \& haze or snow conditions. See \textbf{\emph{supplementary material}} for results of more weather condition and downstream tasks. }
    \label{fig: visual_expand}
\end{figure*}

\paragraph{Analysis on Prior Learning.} As shown in Table~\ref{tab: diffusion}, we show the importance of prior for restoration and the approach of prior learning. First, without the support of priors, reconstruction accuracy drops significantly, resulting in a PSNR loss of approximately 0.64 dB. Secondly, our strategy of using diffusion to learn priors outperforms CNN-based encoders for prior learning by 0.42 dB in PSNR. Finally, unlike traditional diffusion models that estimate feature maps or full images, our method estimates vectors in a one-dimensional latent space. It enables the joint training of diffusion models for prior estimation alongside the prior-based SR network, leading to more accurate results with a PSNR gain of over 1.3 dB compared to diffusion models that estimate full images.

\begin{table}[t!]
\tiny
\setlength\tabcolsep{3pt}
\centering
\resizebox{0.48\textwidth}{!}{
\begin{tabular}{l|cccc|ll}
\toprule
& & & \multicolumn{2}{c|}{Diffusion Space} & \multicolumn{2}{c}{Raindrop} 
\\ 
    \multicolumn{1}{c|}{\multirow{-2}{*}{Params}}
    & \multicolumn{1}{c}{\multirow{-2}{*}{Prior}}
    & \multicolumn{1}{c}{\multirow{-2}{*}{Diffusion}}
    & Latent Space   & Feature Maps
    & PSNR$\uparrow$  & SSIM$\uparrow$  
    \\ 
    \hline
    9.8M & \XSolidBrush &\textcolor{gray}{\XSolidBrush} &\textcolor{gray}{\XSolidBrush} &\textcolor{gray}{\XSolidBrush}  & 25.41 & 0.7397 \\
    9.7M & \textcolor{gray}{\Checkmark} &\XSolidBrush &\textcolor{gray}{\XSolidBrush} &\textcolor{gray}{\XSolidBrush}  & 25.63 & 0.7587 \\
    10.1M & \textcolor{gray}{\Checkmark} &\textcolor{gray}{\Checkmark} &\XSolidBrush &\textcolor{gray}{\Checkmark}  & 24.73 & 0.7459 \\
    \rowcolor{lightgray}
    10.1M & \textcolor{gray}{\Checkmark} &\textcolor{gray}{\Checkmark} &\textcolor{gray}{\Checkmark} 
    &\XSolidBrush  & \textbf{26.05} & \textbf{0.7674} \\

    \toprule
\end{tabular}}
\caption{Ablation on prior learning, where ``Diffusion Space" indicates whether diffusion models learn priors in the latent space or directly from feature maps. \textcolor{gray}{Gray} cells indicate ours.}
\label{tab: diffusion}
\end{table}

\begin{table}[t!]
	\tiny
	\setlength\tabcolsep{2pt}
	\centering
\resizebox{0.48\textwidth}{!}{
\begin{tabular}{l|ccc|ll}
\toprule
&  & & & \multicolumn{2}{c}{Raindrop} 
\\ 
\cmidrule{5-6}
    \multicolumn{1}{c|}{\multirow{-2}{*}{Params}}
    & \multicolumn{1}{c}{\multirow{-2}{*}{Cross-attn}}
    & \multicolumn{1}{c}{\multirow{-2}{*}{Guided Filter}}
    & \multicolumn{1}{c|}{\multirow{-2}{*}{High-pass Filter}}
    & PSNR$\uparrow$  & SSIM$\uparrow$  
    \\ 
    \hline
    8.52M    &\XSolidBrush &\textcolor{gray}{\Checkmark} &\textcolor{gray}{\Checkmark}  & 26.73 & 0.7688 \\
    8.10M     &\textcolor{gray}{\Checkmark} &\XSolidBrush &\textcolor{gray}{\Checkmark}  & 26.48 & 0.7650 \\
    8.10M     &\textcolor{gray}{\Checkmark} 
    &\textcolor{gray}{\Checkmark} &\XSolidBrush  & 26.51 & 0.7660 \\
    7.26M     &\textcolor{gray}{\Checkmark} 
    &\XSolidBrush &\XSolidBrush  & 26.11 & 0.7617 \\
    \rowcolor{lightgray}
    8.94M     &\textcolor{gray}{\Checkmark} &\textcolor{gray}{\Checkmark} &\textcolor{gray}{\Checkmark}  & \textbf{26.82} & \textbf{0.7701} \\

    \toprule
\end{tabular}}
\caption{Ablation on different modules in our method.}
\label{tab: configurations}
\end{table}

\paragraph{Effectiveness of Filters in Our Method.} As shown in Table~\ref{tab: configurations}, we analyze the effect of guided filter (\eg, MR), high-pass filter (\eg, TC), and cross-attention. When priors are selected using either guided or high-pass filters, PSNR increases by an average of 0.33 dB with only a small parameter increase, while removing both results in a performance drop of over 0.7 dB. Cross-attention facilitates fusing post-filter features with input features, bringing a PSNR gain of 0.11 dB. As shown in Fig.\ref{fig: ablation_comparsion}, we further show the role of different modules: guided filters guide priors to remove rain-related media, and high-pass filters help recover sharp textures from high-frequency priors. Fig.\ref{fig: feature_maps} also shows that guided filters aid raindrop removal and high-pass filters aid texture recovery. Combining these visualizations with the quantitative results in Table~\ref{tab: configurations}, \emph{we conclude that these effects appear not only in the pixel domain but also in the feature domain.}

\begin{figure}[t]
	\begin{overpic}[width=0.999\linewidth]{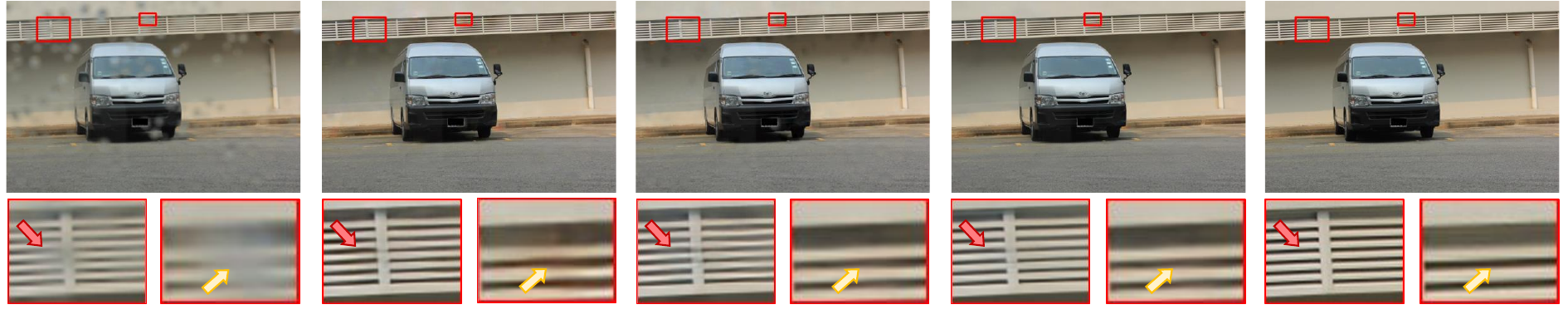}
        \put(4.0,-1.3){\color{black}{\fontsize{7pt}{1pt}\selectfont LR+Rainy}}
		\put(23.1,-1.3){\color{black}{\fontsize{7pt}{1pt}\selectfont w/o Guided}}
		\put(42.0,-1.3){\color{black}{\fontsize{7pt}{1pt}\selectfont w/o High-pass}}
		\put(67.0,-1.3){\color{black}{\fontsize{7pt}{1pt}\selectfont \textbf{Ours}}}
		\put(82.0,-1.3){\color{black}{\fontsize{7pt}{1pt}\selectfont Ground Truth}}	
	\end{overpic}
	\caption{Effects of guided and high-pass filters. Yellow and red arrows indicate artifacts from raindrop removal and blurry edges. Without guided filters, raindrop removal appears as artifacts, but edges remain sharp. Without high-pass filters, edges are blurred, but raindrop removal is relatively clean.}
    \label{fig: ablation_comparsion}
\end{figure}

\begin{figure}[t!]
	\begin{overpic}[width=0.999\linewidth]{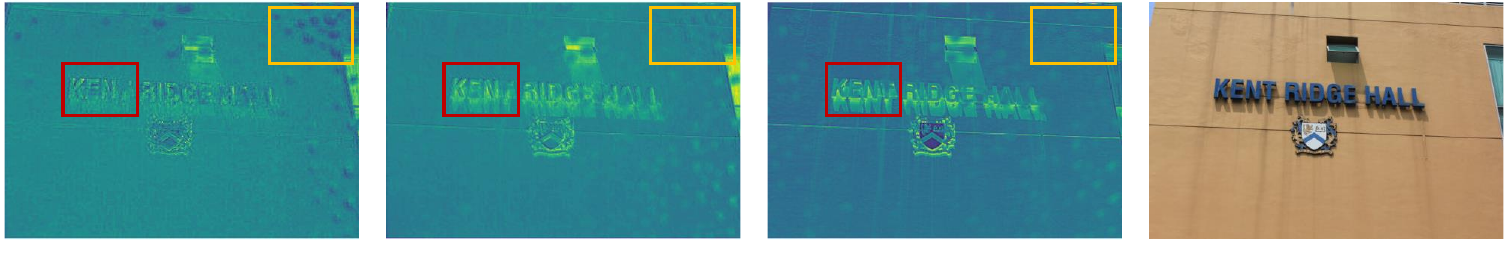}

        \put(5.0,-0.8){\color{black}{\fontsize{7pt}{1pt}\selectfont Baseline}}
		\put(32.7,-0.8){\color{black}{\fontsize{7pt}{1pt}\selectfont +Guided}}
		\put(48.8,-0.8){\color{black}{\fontsize{7pt}{1pt}\selectfont +(Guided \& High-pass)}}
		\put(79.8,-0.8){\color{black}{\fontsize{7pt}{1pt}\selectfont Ground Truth}}
		
	\end{overpic}
	\caption{Feature visualization across different modules shows that guided and high-pass filters still play a role in rain media removal and edge recovery in the feature domain. Yellow and red boxes highlight raindrops and edges.}
    \label{fig: feature_maps}
\end{figure}
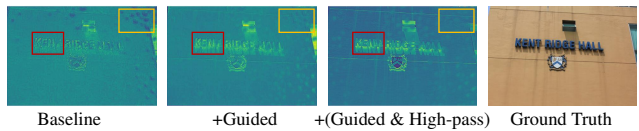

\section{Conclusion}
We introduce DHGM, a diffusion-based high-frequency guided model, which can effectively compensate for the loss of fine textures caused by rain removal and recover potential LR objects under rainy conditions through a unified framework of joint deraining and SR. By fully leveraging pre-trained latent diffusion priors together with guided and high-pass filters, DHGM simultaneously removes complex weather-reduced noise and restores missing high-frequency details. Comprehensive experiments on multiple deraining benchmarks demonstrate that our approach can produces cleaner and HR images than existing deraining, restoration, or cascaded SR pipelines, and significantly improves the perception and detection accuracy of small objects in downstream tasks with less computational cost. 

\section{Acknowledgments}
This work was supported by National Natural Science Foundation of China (Grant No. 62472044, U24B20155, 62225601, U23B2052), Beijing-Tianjin-Hebei Basic Research Funding Program No. F2024502017, Hebei Natural Science Foundation Project No. 242Q0101Z, Beijing Natural Science Foundation Project No. L242025.

\bibliography{aaai2026}

@String(CVPR= {IEEE Conf. Comput. Vis. Pattern Recog.})

@String(ICCV= {Int. Conf. Comput. Vis.})

@String(ECCV= {Eur. Conf. Comput. Vis.})

@String(ICLR = {Int. Conf. Learn. Represent.})

@String(AAAI = {AAAI})

@String(CVPR  = {CVPR})

@String(ICCV  = {ICCV})

@String(ECCV  = {ECCV})

@String(ICLR  = {ICLR})

@article{dong2015image,
  title={Image super-resolution using deep convolutional networks},
  author={Dong, Chao and Loy, Chen Change and He, Kaiming and Tang, Xiaoou},
  journal={IEEE Transactions on Pattern Analysis and Machine Intelligence},
  volume={38},
  number={2},
  pages={295--307},
  year={2015},
  publisher={IEEE}
}

@inproceedings{zhang2018image,
  title={Image super-resolution using very deep residual channel attention networks},
  author={Zhang, Yulun and Li, Kunpeng and Li, Kai and Wang, Lichen and Zhong, Bineng and Fu, Yun},
  booktitle={ECCV},
  pages={286--301},
  year={2018}
}

@inproceedings{dai2019second,
  title={Second-order attention network for single image super-resolution},
  author={Dai, Tao and Cai, Jianrui and Zhang, Yongbing and Xia, Shu-Tao and Zhang, Lei},
  booktitle={CVPR},
  pages={11065--11074},
  year={2019}
}

@inproceedings{mei2021image,
  title={Image super-resolution with non-local sparse attention},
  author={Mei, Yiqun and Fan, Yuchen and Zhou, Yuqian},
  booktitle={CVPR},
  pages={3517--3526},
  year={2021}
}

@inproceedings{gao2022feature,
  title={Feature distillation interaction weighting network for lightweight image super-resolution},
  author={Gao, Guangwei and Li, Wenjie and Li, Juncheng and Wu, Fei and Lu, Huimin and Yu, Yi},
  booktitle={AAAI},
  volume={36},
  number={1},
  pages={661--669},
  year={2022}
}

@inproceedings{yu2018wide,
  title={Wide activation for efficient and accurate image super-resolution},
  author={Yu, Jiahui and Fan, Yuchen and Yang, Jianchao and Xu, Ning and Wang, Zhaowen and Wang, Xinchao and Huang, Thomas},
  booktitle={CVPR NTIRE},
  year={2018}
}

@inproceedings{yuan2021tokens,
  title={Tokens-to-token vit: Training vision transformers from scratch on imagenet},
  author={Yuan, Li and Chen, Yunpeng and Wang, Tao and Yu, Weihao and Shi, Yujun and Jiang, Zi-Hang and Tay, Francis EH and Feng, Jiashi and Yan, Shuicheng},
  booktitle={ICCV},
  pages={558--567},
  year={2021}
}

@inproceedings{chen2021pre,
  title={Pre-trained image processing transformer},
  author={Chen, Hanting and Wang, Yunhe and Guo, Tianyu and Xu, Chang and Deng, Yiping and Liu, Zhenhua and Ma, Siwei and Xu, Chunjing and Xu, Chao and Gao, Wen},
  booktitle={CVPR},
  pages={12299--12310},
  year={2021}
}

@inproceedings{liang2021swinir,
  title={Swinir: Image restoration using swin transformer},
  author={Liang, Jingyun and Cao, Jiezhang and Sun, Guolei and Zhang, Kai and Van Gool, Luc and Timofte, Radu},
  booktitle={ICCVW},
  pages={1833--1844},
  year={2021}
}

@article{li2023cross,
  title={Cross-receptive focused inference network for lightweight image super-resolution},
  author={Li, Wenjie and Li, Juncheng and Gao, Guangwei and Deng, Weihong and Zhou, Jiantao and Yang, Jian and Qi, Guo-Jun},
  journal={IEEE Transactions on Multimedia},
  volume={26},
  pages={864--877},
  year={2023},
  publisher={IEEE}
}

@inproceedings{wang2023omni,
  title={Omni aggregation networks for lightweight image super-resolution},
  author={Wang, Hang and Chen, Xuanhong and Ni, Bingbing and Liu, Yutian and Liu, Jinfan},
  booktitle={CVPR},
  pages={22378--22387},
  year={2023}
}

@inproceedings{zhang2024transcending,
  title={Transcending the limit of local window: Advanced super-resolution transformer with adaptive token dictionary},
  author={Zhang, Leheng and Li, Yawei and Zhou, Xingyu and Zhao, Xiaorui and Gu, Shuhang},
  booktitle={CVPR},
  pages={2856--2865},
  year={2024}
}

@inproceedings{zhou2023srformer,
  title={Srformer: Permuted self-attention for single image super-resolution},
  author={Zhou, Yupeng and Li, Zhen and Guo, Chun-Le and Bai, Song and Cheng, Ming-Ming and Hou, Qibin},
  booktitle={ICCV},
  pages={12780--12791},
  year={2023}
}

@inproceedings{li2020all,
  title={All in one bad weather removal using architectural search},
  author={Li, Ruoteng and Tan, Robby T and Cheong, Loong-Fah},
  booktitle={CVPR},
  pages={3175--3185},
  year={2020}
}

@inproceedings{valanarasu2022transweather,
  title={Transweather: Transformer-based restoration of images degraded by adverse weather conditions},
  author={Valanarasu, Jeya Maria Jose and Yasarla, Rajeev and Patel, Vishal M},
  booktitle={CVPR},
  pages={2353--2363},
  year={2022}
}

@inproceedings{zhang2023weatherstream,
  title={Weatherstream: Light transport automation of single image deweathering},
  author={Zhang, Howard and Ba, Yunhao and Yang, Ethan and Mehra, Varan and Gella, Blake and Suzuki, Akira and Pfahnl, Arnold and Chandrappa, Chethan Chinder and Wong, Alex and Kadambi, Achuta},
  booktitle={CVPR},
  pages={13499--13509},
  year={2023}
}

@inproceedings{zhu2023learning,
  title={Learning weather-general and weather-specific features for image restoration under multiple adverse weather conditions},
  author={Zhu, Yurui and Wang, Tianyu and Fu, Xueyang and Yang, Xuanyu and Guo, Xin and Dai, Jifeng and Qiao, Yu and Hu, Xiaowei},
  booktitle={CVPR},
  pages={21747--21758},
  year={2023}
}

@article{ozdenizci2023restoring,
  title={Restoring vision in adverse weather conditions with patch-based denoising diffusion models},
  author={{\"O}zdenizci, Ozan and Legenstein, Robert},
  journal={IEEE Transactions on Pattern Analysis and Machine Intelligence},
  volume={45},
  number={8},
  pages={10346--10357},
  year={2023},
  publisher={IEEE}
}

@inproceedings{ho2020denoising,
  title={Denoising diffusion probabilistic models},
  author={Ho, Jonathan and Jain, Ajay and Abbeel, Pieter},
  booktitle={NeurIPs},
  volume={33},
  pages={6840--6851},
  year={2020}
}

@inproceedings{patil2023multi,
  title={Multi-weather image restoration via domain translation},
  author={Patil, Prashant W and Gupta, Sunil and Rana, Santu and Venkatesh, Svetha and Murala, Subrahmanyam},
  booktitle={ICCV},
  pages={21696--21705},
  year={2023}
}

@article{tan2024exploring,
  title={Exploring the Application of Large-Scale Pre-Trained Models on Adverse Weather Removal},
  author={Tan, Zhentao and Wu, Yue and Liu, Qiankun and Chu, Qi and Lu, Le and Ye, Jieping and Yu, Nenghai},
  journal={IEEE Transactions on Image Processing},
  year={2024},
  publisher={IEEE}
}

@inproceedings{chen2022learning,
  title={Learning multiple adverse weather removal via two-stage knowledge learning and multi-contrastive regularization: Toward a unified model},
  author={Chen, Wei-Ting and Huang, Zhi-Kai and Tsai, Cheng-Che and Yang, Hao-Hsiang and Ding, Jian-Jiun and Kuo, Sy-Yen},
  booktitle={CVPR},
  pages={17653--17662},
  year={2022}
}

@inproceedings{ye2023adverse,
  title={Adverse weather removal with codebook priors},
  author={Ye, Tian and Chen, Sixiang and Bai, Jinbin and Shi, Jun and Xue, Chenghao and Jiang, Jingxia and Yin, Junjie and Chen, Erkang and Liu, Yun},
  booktitle={ICCV},
  pages={12653--12664},
  year={2023}
}

@inproceedings{park2023all,
  title={All-in-one image restoration for unknown degradations using adaptive discriminative filters for specific degradations},
  author={Park, Dongwon and Lee, Byung Hyun and Chun, Se Young},
  booktitle={CVPR},
  pages={5815--5824},
  year={2023},
}

@inproceedings{sun2024restoring,
  title={Restoring Images in Adverse Weather Conditions via Histogram Transformer},
  author={Sun, Shangquan and Ren, Wenqi and Gao, Xinwei and Wang, Rui and Cao, Xiaochun},
  booktitle={ECCV},
  year={2024}
}

@inproceedings{qian2018attentive,
  title={Attentive generative adversarial network for raindrop removal from a single image},
  author={Qian, Rui and Tan, Robby T and Yang, Wenhan and Su, Jiajun and Liu, Jiaying},
  booktitle={CVPR},
  pages={2482--2491},
  year={2018}
}

@inproceedings{xia2023diffir,
  title={Diffir: Efficient diffusion model for image restoration},
  author={Xia, Bin and Zhang, Yulun and Wang, Shiyin and Wang, Yitong and Wu, Xinglong and Tian, Yapeng and Yang, Wenming and Van Gool, Luc},
  booktitle={ICCV},
  pages={13095--13105},
  year={2023}
}

@inproceedings{quan2021removing,
  title={Removing raindrops and rain streaks in one go},
  author={Quan, Ruijie and Yu, Xin and Liang, Yuanzhi and Yang, Yi},
  booktitle={CVPR},
  pages={9147--9156},
  year={2021}
}

@article{he2012guided,
  title={Guided image filtering},
  author={He, Kaiming and Sun, Jian and Tang, Xiaoou},
  journal={IEEE Transactions on Pattern Analysis and Machine Intelligence},
  volume={35},
  number={6},
  pages={1397--1409},
  year={2012},
  publisher={IEEE}
}

@article{wang2002universal,
  title={A universal image quality index},
  author={Wang, Zhou and Bovik, Alan C},
  journal={IEEE Signal Processing Letters},
  volume={9},
  number={3},
  pages={81--84},
  year={2002},
  publisher={IEEE}
}

@inproceedings{zhang2018unreasonable,
  title={The unreasonable effectiveness of deep features as a perceptual metric},
  author={Zhang, Richard and Isola, Phillip and Efros, Alexei A and Shechtman, Eli and Wang, Oliver},
  booktitle={CVPR},
  pages={586--595},
  year={2018}
}

@article{mittal2012making,
  title={Making a “completely blind” image quality analyzer},
  author={Mittal, Anish and Soundararajan, Rajiv and Bovik, Alan C},
  journal={IEEE Signal Processing Letters},
  volume={20},
  number={3},
  pages={209--212},
  year={2012},
  publisher={IEEE}
}

@article{ding2020image,
  title={Image quality assessment: Unifying structure and texture similarity},
  author={Ding, Keyan and Ma, Kede and Wang, Shiqi and Simoncelli, Eero P},
  journal={IEEE Transactions on Pattern Analysis and Machine Intelligence},
  volume={44},
  number={5},
  pages={2567--2581},
  year={2020},
  publisher={IEEE}
}

@inproceedings{chen2024bidirectional,
  title={Bidirectional multi-scale implicit neural representations for image deraining},
  author={Chen, Xiang and Pan, Jinshan and Dong, Jiangxin},
  booktitle={CVPR},
  pages={25627--25636},
  year={2024}
}

@article{xiao2022image,
  title={Image de-raining transformer},
  author={Xiao, Jie and Fu, Xueyang and Liu, Aiping and Wu, Feng and Zha, Zheng-Jun},
  journal={IEEE Transactions on Pattern Analysis and Machine Intelligence},
  volume={45},
  number={11},
  pages={12978--12995},
  year={2022},
  publisher={IEEE}
}

@inproceedings{jiang2020multi,
  title={Multi-scale progressive fusion network for single image deraining},
  author={Jiang, Kui and Wang, Zhongyuan and Yi, Peng and Chen, Chen and Huang, Baojin and Luo, Yimin and Ma, Jiayi and Jiang, Junjun},
  booktitle={CVPR},
  pages={8346--8355},
  year={2020}
}

@inproceedings{wang2020rethinking,
  title={Rethinking image deraining via rain streaks and vapors},
  author={Wang, Yinglong and Song, Yibing and Ma, Chao and Zeng, Bing},
  booktitle={ECCV},
  pages={367--382},
  year={2020},
  organization={Springer}
}

@inproceedings{rombach2022high,
  title={High-resolution image synthesis with latent diffusion models},
  author={Rombach, Robin and Blattmann, Andreas and Lorenz, Dominik and Esser, Patrick and Ommer, Bj{\"o}rn},
  booktitle={CVPR},
  pages={10684--10695},
  year={2022}
}

@article{uddin2022real,
  title={Real-World Single Image Super-Resolution Under Rainy Condition},
  author={Uddin, Mohammad Shahab},
  journal={arXiv preprint arXiv:2206.08345},
  year={2022}
}

@inproceedings{guo2024onerestore,
  title={Onerestore: A universal restoration framework for composite degradation},
  author={Guo, Yu and Gao, Yuan and Lu, Yuxu and Zhu, Huilin and Liu, Ryan Wen and He, Shengfeng},
  booktitle={ECCV},
  pages={255--272},
  year={2024},
  organization={Springer}
}

@article{li2024efficient,
  title={Efficient image super-resolution with feature interaction weighted hybrid network},
  author={Li, Wenjie and Li, Juncheng and Gao, Guangwei and Deng, Weihong and Yang, Jian and Qi, Guo-Jun and Lin, Chia-Wen},
  journal={IEEE Transactions on Multimedia},
  year={2024},
  publisher={IEEE}
}

@article{khayam2003discrete,
  title={The discrete cosine transform (DCT): theory and application},
  author={Khayam, Syed Ali},
  journal={Michigan State University},
  volume={114},
  number={1},
  pages={31},
  year={2003}
}

@inproceedings{song2020denoising,
  title={Denoising diffusion implicit models},
  author={Song, Jiaming and Meng, Chenlin and Ermon, Stefano},
  booktitle={ICLR},
  year={2020}
}

@article{khan2016importance,
  title={Importance of high order high pass and low pass filters},
  author={Khan, MDNH and Alam, MDM and Masud, MDA and Amin, AA},
  journal={World Applied Sciences Journal},
  volume={34},
  number={9},
  pages={1261--1268},
  year={2016}
}

@inproceedings{karras2019style,
  title={A style-based generator architecture for generative adversarial networks},
  author={Karras, Tero and Laine, Samuli and Aila, Timo},
  booktitle={CVPR},
  pages={4401--4410},
  year={2019}
}

@inproceedings{esser2021taming,
  title={Taming transformers for high-resolution image synthesis},
  author={Esser, Patrick and Rombach, Robin and Ommer, Bjorn},
  booktitle={CVPR},
  pages={12873--12883},
  year={2021}
}

@inproceedings{varghese2024yolov8,
  title={Yolov8: A novel object detection algorithm with enhanced performance and robustness},
  author={Varghese, Rejin and Sambath, M},
  booktitle={ADICS},
  pages={1--6},
  year={2024},
  organization={IEEE}
}

@inproceedings{yang2017deep,
  title={Deep joint rain detection and removal from a single image},
  author={Yang, Wenhan and Tan, Robby T and Feng, Jiashi and Liu, Jiaying and Guo, Zongming and Yan, Shuicheng},
  booktitle={CVPR},
  pages={1357--1366},
  year={2017}
}

@article{jin2020ai,
  title={AI-GAN: Asynchronous interactive generative adversarial network for single image rain removal},
  author={Jin, Xin and Chen, Zhibo and Li, Weiping},
  journal={Pattern Recognition},
  volume={100},
  pages={107143},
  year={2020},
  publisher={Elsevier}
}

@inproceedings{li2025self,
  title={Self-Supervised Selective-Guided Diffusion Model for Old-Photo Face Restoration},
  author={Li, Wenjie and Wang, Xiangyi and Guo, Heng and Gao, Guangwei and Ma, Zhanyu},
  booktitle={NeurIPs},
  year={2025}
}

@article{li2025dual,
  title={Dual-domain modulation network for lightweight image super-resolution},
  author={Li, Wenjie and Guo, Heng and Hou, Yuefeng and Gao, Guangwei and Ma, Zhanyu},
  journal={IEEE Transactions on Multimedia},
  year={2025}
}

@article{li2023survey,
  title={Survey on deep face restoration: From non-blind to blind and beyond},
  author={Li, Wenjie and Wang, Mei and Zhang, Kai and Li, Juncheng and Li, Xiaoming and Zhang, Yuhang and Gao, Guangwei and Deng, Weihong and Lin, Chia-Wen},
  journal={arXiv preprint arXiv:2309.15490},
  year={2023}
}

@inproceedings{li2024efficientface,
  title={Efficient face super-resolution via wavelet-based feature enhancement network},
  author={Li, Wenjie and Guo, Heng and Liu, Xuannan and Liang, Kongming and Hu, Jiani and Ma, Zhanyu and Guo, Jun},
  booktitle={ACM MM},
  pages={4515--4523},
  year={2024}
}

@inproceedings{peng2025boosting,
  title={Boosting image de-raining via central-surrounding synergistic convolution},
  author={Peng, Long and Wang, Yang and Di, Xin and Fu, Xueyang and Cao, Yang and Zha, Zheng-Jun and others},
  booktitle={AAAI},
  volume={39},
  number={6},
  pages={6470--6478},
  year={2025}
}

@article{peng2024lightweight,
  title={Lightweight adaptive feature de-drifting for compressed image classification},
  author={Peng, Long and Cao, Yang and Sun, Yuejin and Wang, Yang},
  journal={IEEE Transactions on Multimedia},
  volume={26},
  pages={6424--6436},
  year={2024},
  publisher={IEEE}
}

@inproceedings{chen2023sparse,
  title={Sparse sampling transformer with uncertainty-driven ranking for unified removal of raindrops and rain streaks},
  author={Chen, Sixiang and Ye, Tian and Bai, Jinbin and Chen, Erkang and Shi, Jun and Zhu, Lei},
  booktitle={ICCV},
  pages={13106--13117},
  year={2023}
}

@inproceedings{pengtowards,
  title={Towards Realistic Data Generation for Real-World Super-Resolution},
  author={Peng, Long and Li, Wenbo and Pei, Renjing and Ren, Jingjing and Xu, Jiaqi and Wang, Yang and Cao, Yang and Zha, Zheng-Jun},
  booktitle={ICLR},
  year={2025}
}

\end{document}